\documentclass{article}

\usepackage{arxiv}

\usepackage[utf8]{inputenc}
\usepackage[T1]{fontenc}
\usepackage[hidelinks]{hyperref}
\usepackage{url}
\usepackage{booktabs}
\usepackage{amsfonts}
\usepackage{amsmath}
\usepackage{amssymb}
\usepackage{nicefrac}
\usepackage{microtype}
\usepackage[capitalize]{cleveref}
\usepackage{graphicx}
\usepackage{natbib}
\usepackage{doi}
\usepackage[caption=false]{subfig}
\usepackage{pifont}
\usepackage{dsfont}
\usepackage{float}
\usepackage[section]{placeins}
\usepackage{mathtools}
\usepackage{epigraph}
\usepackage{siunitx}
\usepackage{needspace}
\usepackage{tikz}
\usetikzlibrary{tikzmark,decorations.pathreplacing,calc}
\usepackage{multirow}
\usepackage{numprint}
\usepackage{thmtools}
\usepackage{thm-restate}
\usepackage{authblk}

\npdecimalsign{.}
\makeatletter
\let\originalfnsymbol\@fnsymbol
\renewcommand\@fnsymbol[1]{\ifcase#1\or\TextOrMath\textdagger\dagger\or\TextOrMath\textdaggerdbl\ddagger\else\originalfnsymbol{#1}\fi}
\makeatother

\title{Partner-aware Peptide--Protein Interaction Prediction and Target-conditioned Peptide Generation}
\date{}

\author[1]{Chupei Tang\thanks{These authors contributed equally as co-first authors.}}
\author[1]{Junxiao Kong\textsuperscript{\textdagger}}
\author[1]{Moyu Tang}
\author[2]{Di Wang}
\author[1,4]{Jixiu Zhai}
\author[1]{Ronghao Xie}
\author[1]{Shangkun Sima}
\author[1,3]{Tianchi Lu\thanks{Correspondence should be addressed to \texttt{tianchilu4-c@my.cityu.edu.hk}; Tel: +86-13239620274.}}

\affil[1]{School of Mathematics and Statistics, Lanzhou University, 222 South Tianshui Road, Lanzhou 730000, Gansu, China}
\affil[2]{Cuiying Honors College, Lanzhou University, Lanzhou, Gansu, China}
\affil[3]{Department of Computer Science, City University of Hong Kong, 83 Tat Chee Avenue, Kowloon Tong, Hong Kong 999077, China}
\affil[4]{Shanghai Innovation Institute, Shanghai, China}

\hypersetup{
pdftitle={Partner-aware Peptide--Protein Interaction Prediction and Target-conditioned Peptide Generation},
pdfsubject={q-bio.BM, q-bio.QM},
pdfauthor={Chupei Tang, Junxiao Kong, Moyu Tang, Di Wang, Jixiu Zhai, Ronghao Xie, Shangkun Sima, Tianchi Lu},
pdfkeywords={peptide-protein interaction, target-conditioned generation, peptide design, protein language model},
hypertexnames=false,
}

\begin{document}
\maketitle
\setlength{\headheight}{24pt}

\begin{abstract}
\textbf{Motivation:} Peptide-protein interactions (PepPIs) are central to cellular regulation and peptide therapeutics, but experimental characterization remains too slow for large-scale screening. Existing methods usually emphasize either interaction prediction or peptide generation, leaving candidate prioritization, residue-level interpretation, and target-conditioned expansion insufficiently integrated.

\textbf{Results:} We present an integrated framework for early-stage peptide screening that combines a partner-aware prediction and localization model (ConGA-PepPI) with a target-conditioned generative model (TC-PepGen). ConGA-PepPI uses asymmetric encoding, bidirectional cross-attention, and progressive transfer from pair prediction to binding-site localization, while TC-PepGen preserves target information throughout autoregressive decoding via layerwise conditioning. In five-fold cross-validation, ConGA-PepPI achieved 0.839 accuracy and 0.921 AUROC, with binding-site AUPR values of 0.601 on the protein side and 0.950 on the peptide side, and remained competitive on external benchmarks. Under a controlled length-conditioned benchmark, 40.39\% of TC-PepGen peptides exceeded native templates in AlphaFold 3 ipTM, and unconstrained generation retained evidence of target-conditioned signal.

\end{abstract}


\section{Introduction}\label{sec:introduction}

Peptide-protein interactions (PepPIs) regulate diverse cellular processes and are increasingly important in peptide therapeutics \citep{liu2018sall4,xu2024lysineStapled}. Experimental characterization remains slow and labor-intensive, limiting large-scale screening and motivating computational approaches for interaction prediction and peptide design. There is therefore a need for computational frameworks that can both prioritize likely binders and expand candidate sequence space while preserving the biological specificity of individual peptide-protein pairs.

Existing computational strategies for PepPI analysis broadly fall into three categories: structure-based prediction, sequence-based prediction, and target-conditioned generation. Structure-based methods can produce detailed binding hypotheses but depend on resolved or modeled three-dimensional structures and remain difficult to scale, whereas sequence-based prediction and recent target-conditioned generation methods are more practical for large datasets \citep{kurcinski2015cabsdock,xu2018mdockpep,zhang2019autodockcrankpep,lee2015galaxypepdock,lei2021camp,wu2022bridgedpi,wang2024deeppeppi,chen2025plpa,chen2025targetSequence}. What remains less developed is a unified approach that links pair prioritization and residue localization with target-conditioned candidate expansion for early-stage computational screening. The key challenge is therefore not isolated prediction or generation, but connecting these capabilities in a target-specific and interpretable screening setting.

Despite this progress, several practical limitations remain on the prediction side. Peptides and proteins are intrinsically asymmetric interaction partners: peptides are short and often governed by localized binding motifs, whereas proteins are longer and exhibit broader contextual dependencies. This asymmetry suggests that a uniform encoder may not optimally represent both sides of the interaction. In addition, independently encoded sequences followed by late fusion may miss partner-specific interaction context, even though PepPI recognition is inherently conditional on the paired protein sequence. A further challenge is that sequence-level interaction labels are much easier to collect than residue-level binding annotations, making it difficult to learn residue localization directly from limited supervision.

The generation side presents a related but distinct challenge. The task is not generic peptide language modeling, but target-conditioned design in which protein-specific information must remain available as the peptide sequence grows. Conditioning that is injected only once can be diluted during autoregressive decoding, so the model must preserve target awareness throughout generation.

Together, the two components support such a screening-oriented framework for peptide candidate triage. ConGA-PepPI improves binary interaction prediction while also supporting paired binding-site identification, whereas TC-PepGen expands candidate sequence space under persistent target-protein conditioning.

\section{Materials and methods}\label{sec:methods}

\begin{figure}[!t]
	\centering
	\includegraphics[width=\textwidth,height=0.48\textheight,keepaspectratio]{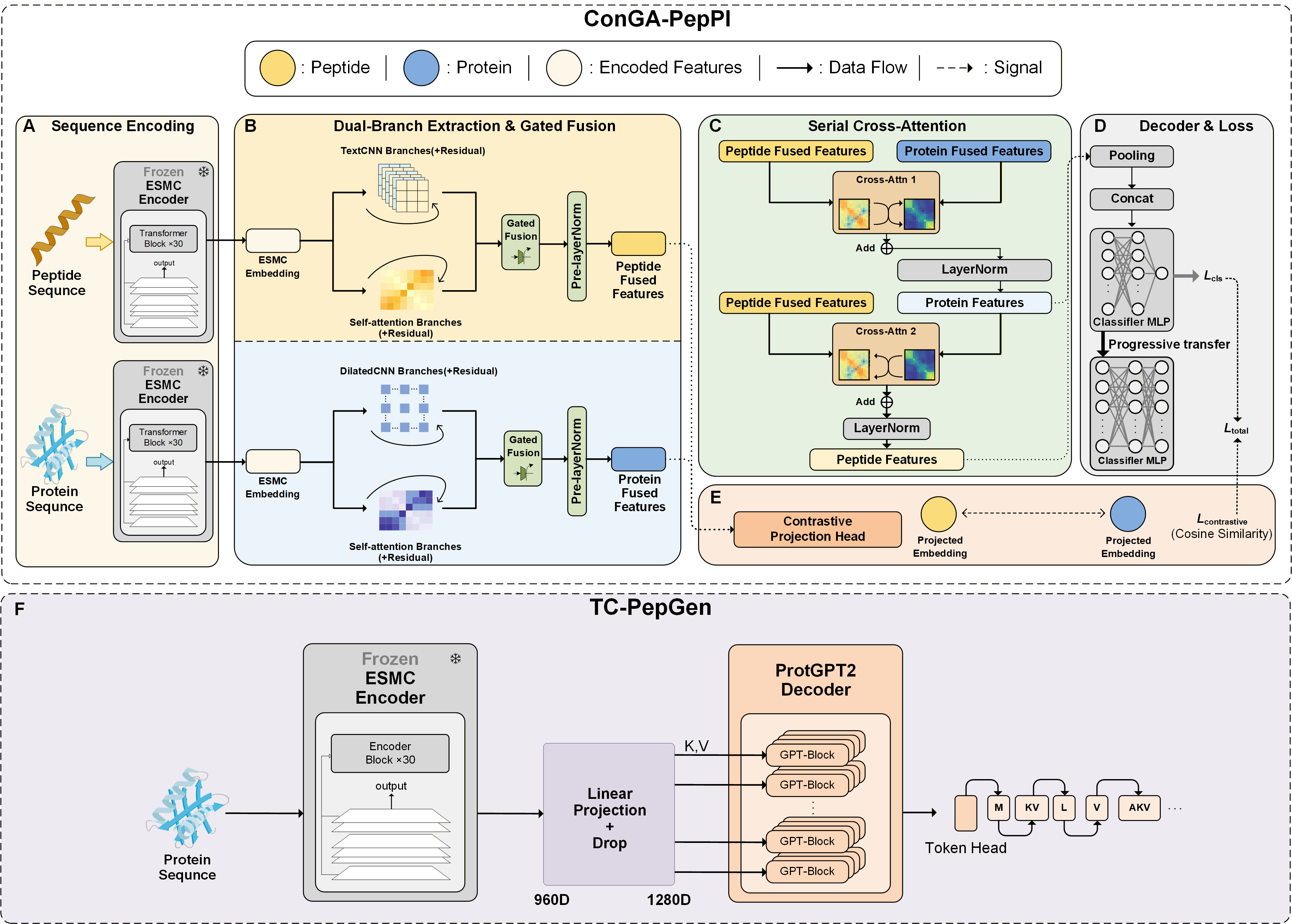}
	\caption{Overview of the ConGA-PepPI and TC-PepGen framework. (A) Peptide and protein sequences are encoded by a frozen ESMC encoder. (B) ConGA-PepPI uses asymmetric dual-branch extraction and gated fusion to capture local and global sequence features. (C) Serial cross-attention refines pair-specific peptide--protein representations. (D) The decoder supports sequence-level interaction prediction and residue-level binding-site identification through progressive transfer learning. (E) A supervised contrastive head improves representation separability. (F) TC-PepGen performs target-conditioned peptide generation through cross-attention in a ProtGPT2 decoder.}
	\label{fig:framework}
\end{figure}

We develop an integrated framework for peptide-protein interaction modeling and target-conditioned peptide generation. It contains two linked components: ConGA-PepPI for candidate prioritization and residue localization, and TC-PepGen for target-conditioned candidate expansion (Figure~\ref{fig:framework}). In this section, we focus on the data sources, preprocessing strategy, and model design principles rather than implementation-level details.

\subsection{Data Acquisition and Preprocessing}

Task-specific datasets were constructed separately for prediction and generation.

For prediction, peptide-protein complexes released before 2023 were collected from the RCSB PDB database, and residue-level interface annotations were extracted with PDB-BRE. Protein chains were aligned to UniProt so that sequence records and annotations could be standardized across samples. Sequence pairs containing more than 20\% unknown or non-standard amino acids were removed. For model input consistency, peptide sequences were truncated or padded to length 50 and protein sequences were truncated or padded to length 800.

Because experimentally verified non-binders are rarely available for PepPI tasks, negative samples were approximated by randomly pairing peptides with non-cognate proteins at a 1:1 positive-to-negative ratio. We used five-fold random splitting for cross-validation. We also constructed an external Test167 benchmark from more recently released complexes after removing pairs with sequence similarity above 80\% to the training or validation sets, allowing us to evaluate both in-distribution performance and stricter low-homology generalization.

For generation, we merged the peptide-target dataset released with PepNN with Propedia and removed duplicate entries. Target proteins were clustered with MMseqs2 at 0.8 sequence identity, and repeated peptide-target combinations within the same cluster were consolidated to one representative entry. We further removed entries whose target proteins exceeded 80\% similarity to preserve a stricter generalization setting. This procedure yielded 10,203 unique peptide-protein pairs, of which 10,000 were used for training and 203 were reserved as a held-out generation test set. To keep the generation task within a consistent biological and computational range, binding peptides were limited to at most 50 residues and target proteins to at most 500 residues.

\subsection{Sequence Representation}

We use a frozen ESMC protein language model as the shared sequence encoder in both ConGA-PepPI and TC-PepGen. In the prediction task, peptide and protein sequences are encoded independently and then combined through partner-aware interaction modeling. In the generation task, only the target protein is encoded, and its representation is provided to the decoder as conditioning information during peptide generation.

\subsection{Prediction Model Overview}

ConGA-PepPI is designed around three practical challenges in PepPI prediction: peptides and proteins are asymmetric interaction partners, interaction recognition depends on the paired context rather than isolated sequence features, and residue-level annotations are much scarcer than sequence-level interaction labels.

To address partner asymmetry, ConGA-PepPI uses branch-specific feature extraction for peptides and proteins so that short motif-like peptide signals and longer protein-context signals can be processed differently. Local and global sequence cues are then fused within each branch to form residue-level representations that remain suitable for downstream interpretation.

To address pair dependence, ConGA-PepPI uses bidirectional interaction modeling so that peptide features can refine protein-side representations and protein context can in turn update peptide-side representations. This design encourages the model to represent a peptide-protein pair as a conditional interaction rather than as two unrelated embeddings concatenated at the end.

For prediction and localization, we use a progressive transfer strategy. The model is first trained for sequence-level interaction prediction on the larger pair-level dataset and then adapted to residue-level binding-site identification on the smaller interface-annotation dataset. This allows the residue-level task to benefit from pair-level supervision learned at a much larger scale.

We also apply supervised contrastive regularization during sequence-level training to improve the separability of positive and negative peptide-protein pairs in latent space. In practice, this regularization complements the classifier and improves the quality of pair representations used for both prioritization and downstream localization.

\subsection{Generation Strategies and Length Constraints}

We evaluated TC-PepGen in two settings because they reflect different deployment scenarios. In the length-conditioned setting, generation is constrained to the native binder length so that comparisons with PepMLM and RFdiffusion remain controlled and directly comparable. In the unconstrained setting, the model predicts peptide length freely through normal autoregressive termination, which provides a stricter test of whether target-conditioned signal is retained without external length guidance.

\subsection{Target-Conditioned Generation Model Overview}

TC-PepGen combines a frozen target-protein encoder with an autoregressive peptide decoder. The design goal is not generic peptide language modeling, but target-conditioned generation in which information from the target protein remains available throughout decoding.

To achieve this, target-protein representations are injected into the decoder through cross-attention at each stage of autoregressive generation rather than being used only as a single initial condition. This allows the model to keep referencing target-specific information as the peptide sequence grows, which is important for preserving binding-related specificity instead of drifting toward an unconditional peptide prior.

Under the controlled benchmark, TC-PepGen generates peptides under a fixed target length so that structure-based comparisons with baseline generators are fair. Under unconstrained generation, the same target-conditioned architecture is used without externally supplied peptide length, allowing us to evaluate whether the model can still recover biologically plausible peptide lengths and target-dependent sequence patterns.

\section{Results}\label{sec:results}
\begin{table}[ht]
	\centering
	\scriptsize
	\setlength{\tabcolsep}{4pt}
	\caption{Ablation study configuration.}
	\label{tab:ablation_config}
	\begin{tabular*}{\textwidth}{@{\extracolsep{\fill}}lccccccc}
		\toprule
		\textbf{Model} & \textbf{Base Encoder} & \shortstack{\textbf{Bidirectional}\\\textbf{Cross-Attn}} & \textbf{Self-Attention} & \textbf{Dual-Branch} & \shortstack{\textbf{Dilated CNN}\\\textbf{for Protein}} & \shortstack{\textbf{Gated}\\\textbf{Fusion}} & \shortstack{\textbf{Contrastive}\\\textbf{Learning}} \\
		\midrule
		Model 0        & \ding{51}             & \ding{55}                          & \ding{55} & \ding{55} & \ding{55} & \ding{55} & \ding{55} \\
		Model 1        & \ding{51}             & \ding{51}                          & \ding{55} & \ding{55} & \ding{55} & \ding{55} & \ding{55} \\
		Model 2        & \ding{51}             & \ding{51}                          & \ding{51} & \ding{55} & \ding{55} & \ding{55} & \ding{55} \\
		Model 3        & \ding{51}             & \ding{51}                          & \ding{51} & \ding{51} & \ding{55} & \ding{55} & \ding{55} \\
		Model 4        & \ding{51}             & \ding{51}                          & \ding{51} & \ding{51} & \ding{51} & \ding{55} & \ding{55} \\
		Model 5        & \ding{51}             & \ding{51}                          & \ding{51} & \ding{51} & \ding{51} & \ding{51} & \ding{55} \\
		ConGA-PepPI    & \ding{51}             & \ding{51}                          & \ding{51} & \ding{51} & \ding{51} & \ding{51} & \ding{51} \\
		\bottomrule
	\end{tabular*}
\end{table}

\begin{table}[ht]
	\centering
	\scriptsize
	\setlength{\tabcolsep}{4pt}
	\caption{Ablation performance under five-fold cross-validation.}
	\label{tab:validation_performance}
	\begin{tabular*}{\textwidth}{@{\extracolsep{\fill}}lccccccc}
		\toprule
		\textbf{Model} & \textbf{Precision}       & \textbf{Recall}          & \textbf{MCC}             & \textbf{ACC}             & \textbf{F1}              & \textbf{AUROC}           & \textbf{AUPR}            \\
		\midrule
		Model 0        & 0.796$\pm$0.010          & 0.828$\pm$0.014          & 0.616$\pm$0.010          & 0.808$\pm$0.005          & 0.812$\pm$0.005          & 0.885$\pm$0.006          & 0.880$\pm$0.004          \\
		Model 1        & 0.799$\pm$0.008          & 0.816$\pm$0.010          & 0.611$\pm$0.017          & 0.806$\pm$0.009          & 0.808$\pm$0.008          & 0.887$\pm$0.006          & 0.883$\pm$0.004          \\
		Model 2        & 0.808$\pm$0.006          & \textbf{0.831$\pm$0.007} & 0.633$\pm$0.007          & 0.817$\pm$0.004          & 0.819$\pm$0.002          & 0.891$\pm$0.003          & 0.879$\pm$0.002          \\
		Model 3        & 0.814$\pm$0.003          & 0.829$\pm$0.008          & 0.639$\pm$0.012          & 0.820$\pm$0.006          & 0.821$\pm$0.005          & 0.895$\pm$0.007          & 0.883$\pm$0.007          \\
		Model 4        & 0.816$\pm$0.010          & \textbf{0.831$\pm$0.010} & 0.643$\pm$0.014          & 0.821$\pm$0.007          & 0.823$\pm$0.006          & 0.897$\pm$0.004          & 0.888$\pm$0.002          \\
		Model 5        & 0.822$\pm$0.003          & 0.824$\pm$0.005          & 0.645$\pm$0.011          & 0.823$\pm$0.005          & 0.823$\pm$0.004          & 0.898$\pm$0.004          & 0.890$\pm$0.004          \\
		ConGA-PepPI    & \textbf{0.880$\pm$0.003} & 0.785$\pm$0.009          & \textbf{0.682$\pm$0.010} & \textbf{0.839$\pm$0.005} & \textbf{0.830$\pm$0.006} & \textbf{0.921$\pm$0.003} & \textbf{0.915$\pm$0.002} \\
		\bottomrule
	\end{tabular*}
\end{table}
\subsection{Ablation Analysis of the Prediction Component}

\vspace{0.3em}

To determine whether ConGA-PepPI improves prediction because its components address the bottlenecks defined in the Introduction rather than because it is simply larger, we introduced bidirectional interaction modeling, branch specialization, gated fusion, and supervised contrastive learning step by step from a common encoder-only baseline (Table~\ref{tab:ablation_config}). Table~\ref{tab:validation_performance} shows that performance improves progressively across this ablation series.

The first gains arise from explicitly modeling partner-dependent context. Relative to the encoder-only baseline, adding bidirectional cross-attention improves discrimination, consistent with the view that PepPI recognition depends on the paired sequence rather than on independent sequence encoding alone. Further gains appear when the model is allowed to treat peptides and proteins asymmetrically through branch specialization and the dilated-convolution protein pathway, supporting the design premise that the two interaction partners operate on different length scales and contextual regimes.

The final improvements come from feature integration and regularization. Gated fusion improves performance further, indicating that prediction benefits from adaptively weighting local and contextual features. Supervised contrastive learning then provides the strongest final gain, especially in precision and MCC. The full model achieves the best results, with accuracy increasing from 0.808 to 0.839 and AUROC from 0.885 to 0.921.

\begin{figure}[!htbp]
	\centering
	\includegraphics[width=0.80\textwidth]{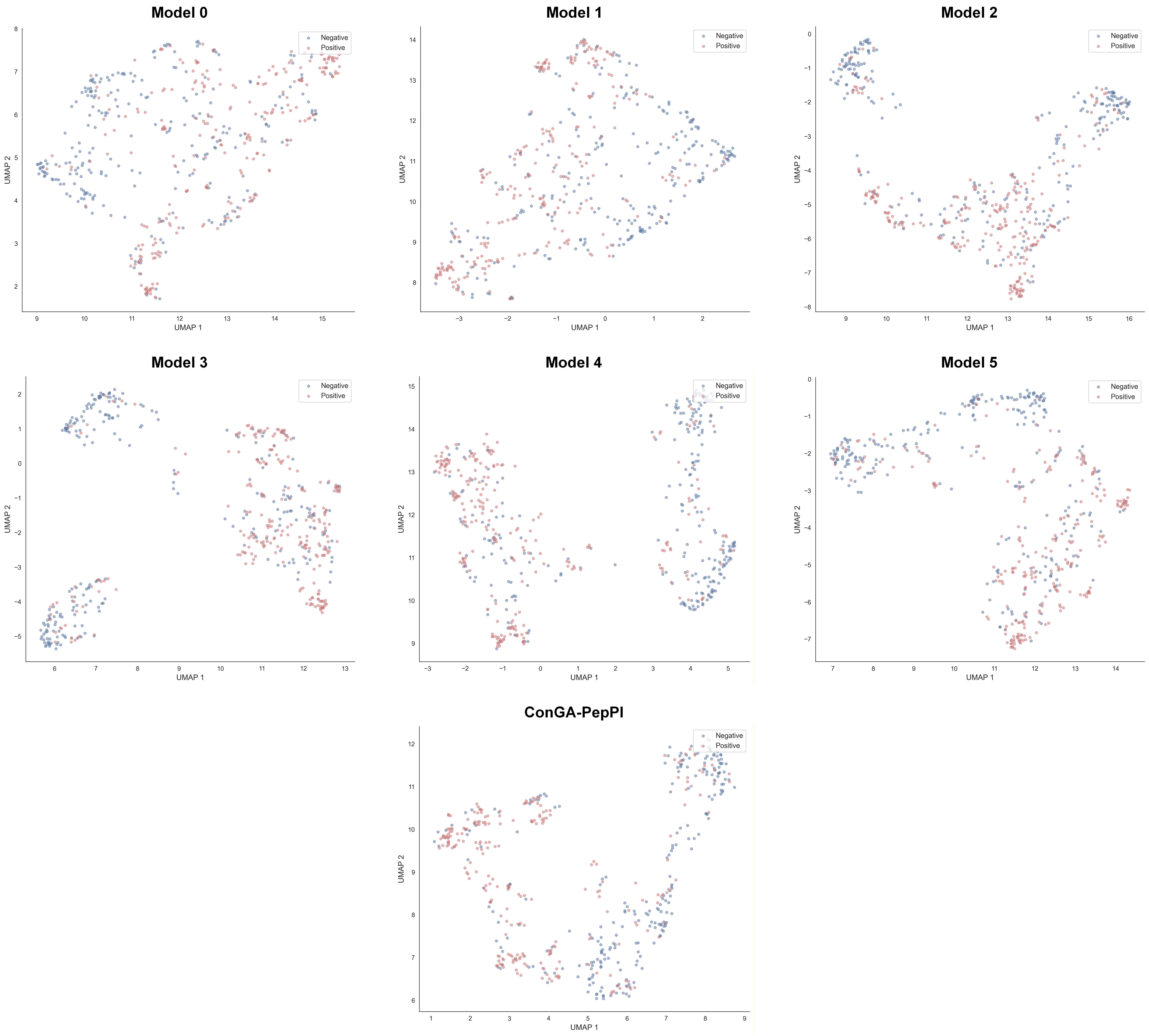}
	\caption{UMAP visualization of feature representations from the ablation series at epoch 50. Red and blue points denote positive and negative peptide--protein pairs, respectively. Panels (a--f) show progressively enriched model variants, starting from the baseline configuration and then adding asymmetric feature extraction, gated fusion, serial interaction modeling, progressive transfer learning, and supervised contrastive learning. As the architecture becomes more complete, the positive and negative samples show progressively clearer separation in latent space, with the clearest class separation observed for the full model.}
	\label{fig:ablation_umap}
\end{figure}

\subsection{Candidate Prioritization through Binary Peptide-Protein Interaction Prediction}

We next asked whether the gains observed in the ablation study translate into stronger end-to-end candidate prioritization against competitive methods. We therefore compared ConGA-PepPI with classical machine-learning baselines and representative recent neural predictors, including CAMP, DeepPepPI, and IIDL-PepPI. As summarized in Table~\ref{tab:method_comparison}, ConGA-PepPI attains the best ACC, precision, F1, MCC, AUROC, and AUPR among the compared methods.

\begin{figure}[!htbp]
	\centering
	\includegraphics[width=0.94\columnwidth]{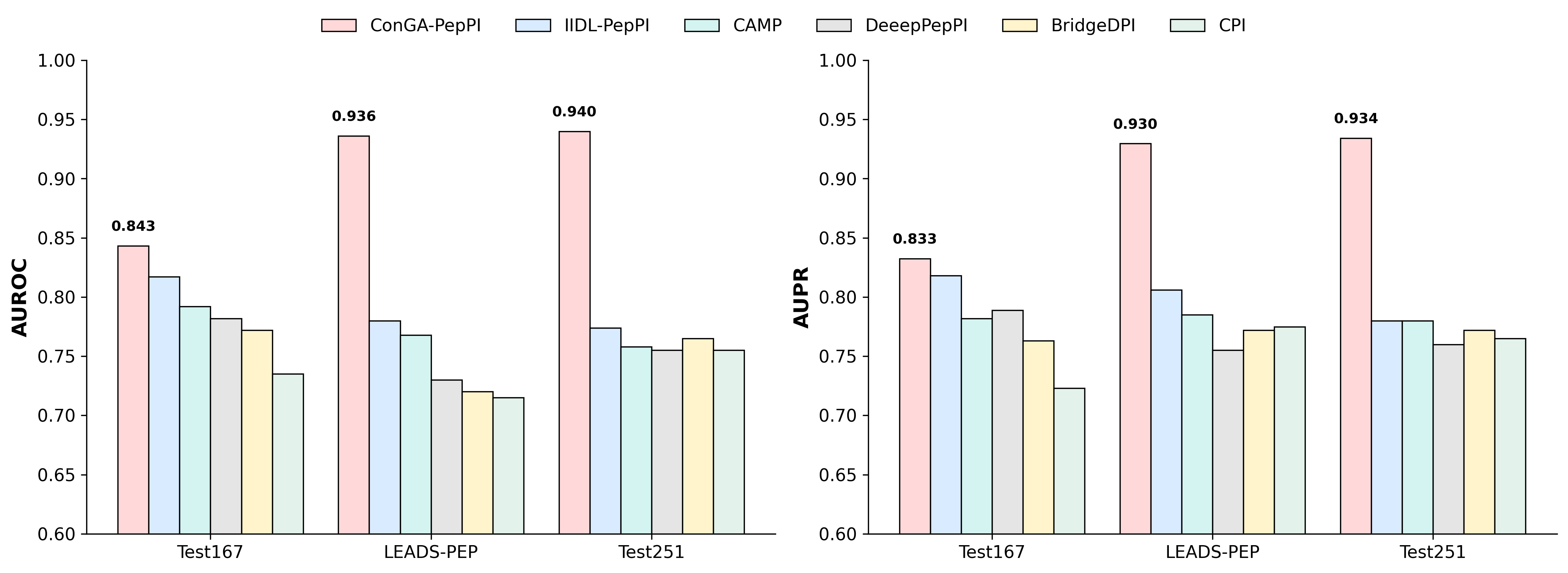}
	\caption{Performance comparison on external benchmark datasets. Left, AUROC; right, AUPR. ConGA-PepPI is compared with IIDL-PepPI, CAMP, DeepPepPI, BridgeDPI, and CPI on Test167, LEADS-PEP, and Test251. The model remains competitive across all three benchmarks.}
	\label{fig:model_comparison}
\end{figure}

We used Test167, a temporally separated low-homology set constructed after removing pairs with sequence similarity above 80\% to the training or validation data, and we further evaluated the model on LEADS-PEP and Test251. Across these three benchmarks, ConGA-PepPI remains competitive relative to representative sequence-based and structure-based baselines (Figure~\ref{fig:model_comparison}), and it achieves an AUROC of 0.936 on LEADS-PEP (Table~\ref{tab:leads_pep_comparison}).

\begin{table}[ht]
	\centering
	\scriptsize
	\setlength{\tabcolsep}{4pt}
	\caption{Method comparison under five-fold cross-validation.}
	\label{tab:method_comparison}
	\begin{tabular*}{\textwidth}{@{\extracolsep{\fill}}lcccccccc}
		\toprule
		\textbf{Method} & \textbf{Precision}       & \textbf{Recall}          & \textbf{MCC}             & \textbf{ACC}             & \textbf{F1}              & \textbf{AUROC}           & \textbf{AUPR}            \\
		\midrule
		LR              & 0.615$\pm$0.046          & 0.563$\pm$0.092          & 0.203$\pm$0.023          & 0.598$\pm$0.007          & 0.579$\pm$0.043          & 0.632$\pm$0.012          & 0.663$\pm$0.012          \\
		SVM             & 0.725$\pm$0.010          & 0.517$\pm$0.011          & 0.335$\pm$0.010          & 0.660$\pm$0.005          & 0.604$\pm$0.007          & 0.760$\pm$0.005          & 0.773$\pm$0.005          \\
		XGBoost         & 0.718$\pm$0.009          & 0.699$\pm$0.008          & 0.425$\pm$0.013          & 0.712$\pm$0.006          & 0.708$\pm$0.007          & 0.783$\pm$0.009          & 0.797$\pm$0.010          \\
		RF              & 0.739$\pm$0.008          & 0.712$\pm$0.007          & 0.461$\pm$0.012          & 0.730$\pm$0.006          & 0.725$\pm$0.006          & 0.800$\pm$0.006          & 0.817$\pm$0.006          \\
		PIPR            & 0.720$\pm$0.010          & 0.717$\pm$0.032          & 0.439$\pm$0.014          & 0.719$\pm$0.007          & 0.718$\pm$0.014          & 0.788$\pm$0.005          & 0.769$\pm$0.006          \\
		DrugBAN       & 0.650$\pm$0.037          & 0.666$\pm$0.062          & 0.305$\pm$0.052          & 0.651$\pm$0.027          & 0.655$\pm$0.028          & 0.715$\pm$0.026          & 0.701$\pm$0.022          \\
		DeepDTA    & 0.728$\pm$0.013          & 0.715$\pm$0.030          & 0.448$\pm$0.010          & 0.723$\pm$0.005          & 0.721$\pm$0.009          & 0.796$\pm$0.006          & 0.777$\pm$0.007          \\
		BridgeDPI    & 0.761$\pm$0.074          & 0.677$\pm$0.057          & 0.473$\pm$0.027          & 0.733$\pm$0.015          & 0.715$\pm$0.032          & 0.822$\pm$0.013          & 0.807$\pm$0.010          \\
		CPI           & 0.771$\pm$0.032          & 0.762$\pm$0.041          & 0.510$\pm$0.013          & 0.754$\pm$0.007          & 0.756$\pm$0.007          & 0.843$\pm$0.007          & 0.825$\pm$0.010          \\
		DeepPepPI  & 0.740$\pm$0.014          & 0.796$\pm$0.022          & 0.519$\pm$0.033          & 0.759$\pm$0.016          & 0.767$\pm$0.016          & 0.835$\pm$0.017          & 0.835$\pm$0.023          \\
		CAMP             & 0.767$\pm$0.015          & 0.775$\pm$0.031          & 0.536$\pm$0.010          & 0.767$\pm$0.005          & 0.769$\pm$0.009          & 0.850$\pm$0.003          & 0.838$\pm$0.006          \\
		IIDL-PepPI      & 0.794$\pm$0.015          & \textbf{0.809$\pm$0.010} & 0.614$\pm$0.014          & 0.807$\pm$0.007          & 0.807$\pm$0.011          & 0.889$\pm$0.005          & 0.882$\pm$0.007          \\
		ConGA-PepPI     & \textbf{0.880$\pm$0.003} & 0.785$\pm$0.009          & \textbf{0.682$\pm$0.010} & \textbf{0.839$\pm$0.005} & \textbf{0.830$\pm$0.006} & \textbf{0.921$\pm$0.003} & \textbf{0.915$\pm$0.002} \\
		\bottomrule
	\end{tabular*}
\end{table}

\begin{table}[ht]
	\centering
	\scriptsize
	\setlength{\tabcolsep}{3pt}
	\caption{Performance of methods on LEADS-PEP.}
	\label{tab:leads_pep_comparison}
	\begin{tabular*}{\columnwidth}{@{\extracolsep{\fill}}lc}
		\toprule
		\textbf{Method}        & \textbf{AUROC} \\
		\midrule
		ConGA-PepPI            & \textbf{0.936} \\
		IIDL-PepPI             & 0.780          \\
		GalaxyPepDock     & 0.773          \\
		AutoDock CrankPep & 0.571          \\
		MDockPeP          & 0.493          \\
		CABS-Dock         & 0.458          \\
		\bottomrule
	\end{tabular*}
\end{table}

\Needspace{8\baselineskip}
\subsection{Residue Localization to Support Candidate Prioritization}

A natural next question is whether the pair-level representation learned by ConGA-PepPI remains useful for residue-level localization when supervision becomes weaker and more spatially specific. To answer this, we initialized the binding-site model from the trained interaction predictor and fine-tuned it on positive samples from each fold. This transfer would complement candidate prioritization with residue-level interpretation.

Protein binding-site prediction is the more difficult side of the task because long proteins and sparse interface residues create severe class imbalance. Even under this setting, ConGA-PepPI achieves the highest AUPR (0.601) and MCC (0.563) among the compared methods (Table~\ref{tab:binding_site_comparison}), indicating that the transferred representation remains informative for residue localization on the protein side.

Peptide binding-site prediction shows an even clearer advantage. On the peptide side, ConGA-PepPI reaches an AUPR of 0.950 and an AUROC of 0.922, together with the best MCC, ACC, F1, and recall among the compared methods (Table~\ref{tab:binding_site_comparison}). Transfer learning mainly improves peptide-side performance, whereas protein-side changes are smaller and metric-dependent (Table~\ref{tab:transfer_learning_comparison}).

\begin{table}[ht]
	\centering
	\scriptsize
	\setlength{\tabcolsep}{4pt}
	\caption{Comparison of methods for protein- and peptide-binding residue identification. ConGA-PepPI performs best overall on the peptide side and achieves the strongest protein-side AUPR and MCC.}
	\label{tab:binding_site_comparison}
	\begin{tabular*}{\textwidth}{@{\extracolsep{\fill}}lcccccccc}
		\toprule
		\textbf{Method} & \textbf{Precision} & \textbf{Recall} & \textbf{MCC}   & \textbf{ACC}   & \textbf{F1}    & \textbf{AUROC} & \textbf{AUPR}  \\
		\midrule
		\multicolumn{8}{c}{\textbf{Protein Binding Residue Identification}}                                                                         \\
		\midrule
		PepBind    & 0.560              & 0.051           & 0.158          & 0.952          & 0.093          & 0.647          & 0.179          \\
		PepNN      & 0.155              & \textbf{0.524}  & 0.218          & 0.836          & 0.239          & 0.755          & 0.190          \\
		TAPE       & 0.599              & 0.026           & 0.117          & 0.957          & 0.049          & 0.775          & 0.219          \\
		ESM-2      & 0.674              & 0.228           & 0.376          & 0.957          & 0.340          & 0.830          & 0.391          \\
		ProtT5      & 0.726              & 0.223           & 0.388          & 0.958          & 0.342          & 0.835          & 0.407          \\
		PepBCL     & 0.519              & 0.262           & 0.347          & 0.952          & 0.348          & 0.625          & 0.409          \\
		AlphaFold3  & 0.594              & 0.497           & 0.528          & 0.969          & 0.541          & 0.783          & 0.544          \\
		PepCA      & \textbf{0.878}     & 0.332           & 0.494          & 0.963          & 0.466          & \textbf{0.889} & 0.549          \\
		IIDL-PepPI      & 0.719              & 0.427           & 0.537          & 0.964          & 0.536          & 0.882          & 0.567          \\
		ConGA-PepPI     & 0.714              & 0.475           & \textbf{0.563} & 0.961          & \textbf{0.570} & 0.882          & \textbf{0.601} \\
		\midrule
		\multicolumn{8}{c}{\textbf{Peptide Binding Residue Identification}}                                                                         \\
		\midrule
		ESM-2      & 0.740              & 0.768           & 0.269          & 0.673          & 0.753          & 0.711          & 0.821          \\
		TAPE       & 0.760              & 0.779           & 0.325          & 0.696          & 0.770          & 0.735          & 0.834          \\
		CAMP        & \textbf{0.905}     & 0.225           & 0.234          & 0.480          & 0.360          & 0.759          & 0.837          \\
		AlphaFold3  & 0.821              & 0.589           & 0.335          & 0.649          & 0.686          & 0.718          & 0.847          \\
		ProtT5      & 0.757              & 0.824           & 0.347          & 0.713          & 0.789          & 0.755          & 0.850          \\
		IIDL-PepPI      & 0.802              & 0.884           & 0.505          & 0.782          & 0.841          & 0.849          & 0.908          \\
		ConGA-PepPI     & 0.882              & \textbf{0.962}  & \textbf{0.724} & \textbf{0.885} & \textbf{0.920} & \textbf{0.922} & \textbf{0.950} \\
		\bottomrule
	\end{tabular*}
\end{table}

\begin{table}[htbp]
	\centering
	\caption{Transfer learning for binding-site prediction. Transfer mainly improves peptide-side localization, while protein-side changes are smaller and metric-dependent.}
	\label{tab:transfer_learning_comparison}
	\resizebox{\columnwidth}{!}{%
		\begin{tabular}{lcccccccc}
			\toprule
			\textbf{Training Strategy} & \textbf{Precision} & \textbf{Recall} & \textbf{MCC}   & \textbf{ACC}   & \textbf{F1}    & \textbf{AUROC} & \textbf{AUPR}  \\
			\midrule
			\multicolumn{8}{c}{\textbf{Peptide Binding Residue Identification}}                                                                                    \\
			\midrule
			Without Transfer Learning  & 0.883              & 0.953           & 0.714          & 0.881          & 0.917          & 0.906          & 0.941          \\
			With Transfer Learning     & 0.882              & \textbf{0.962}  & \textbf{0.724} & \textbf{0.885} & \textbf{0.920} & \textbf{0.922} & \textbf{0.950} \\
			\midrule
			\multicolumn{8}{c}{\textbf{Protein Binding Residue Identification}}                                                                                    \\
			\midrule
			Without Transfer Learning  & 0.698              & 0.488           & 0.564          & 0.961          & 0.574          & 0.881          & 0.594          \\
			With Transfer Learning     & \textbf{0.714}     & 0.475           & 0.563          & \textbf{0.961} & 0.570          & \textbf{0.882} & \textbf{0.601} \\
			\bottomrule
		\end{tabular}%
	}
\end{table}

\subsection{Mechanistic Interpretation of the Prediction Component}
\begin{figure}[!htbp]
	\centering
	\includegraphics[width=0.94\columnwidth]{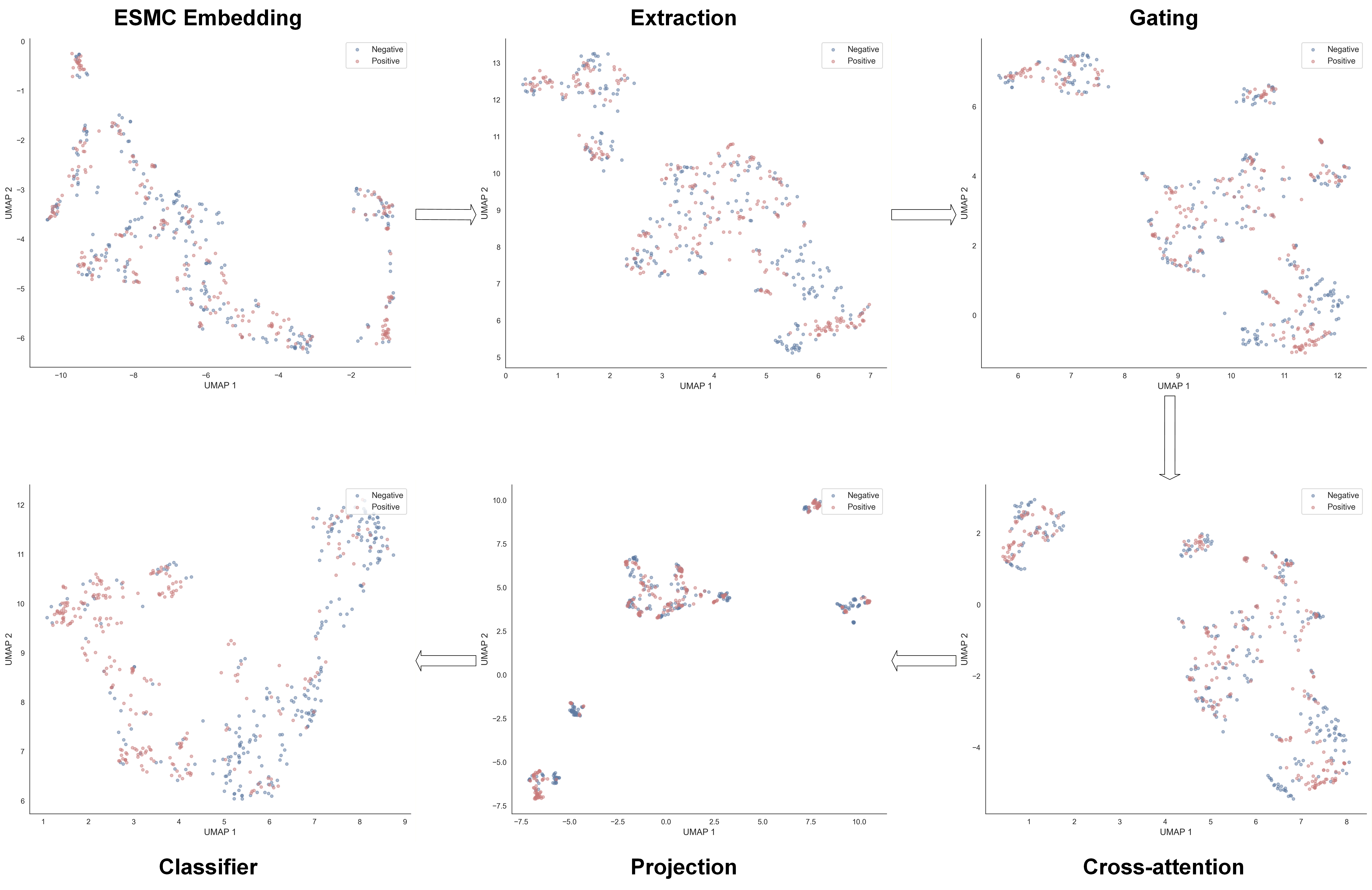}
	\caption{UMAP visualization of intermediate representations from ConGA-PepPI on Test167. Red and blue denote positive and negative peptide--protein pairs, respectively. From top left to bottom right, the panels show ESMC embedding, extraction, gated fusion, cross-attention, projection, and classifier-input features. Class separation becomes progressively clearer through the network.}
	\label{fig:stage6_classifier}
\end{figure}

To understand why ConGA-PepPI improves prioritization rather than merely whether it does, we first visualized classifier-input features from Test167 with UMAP. Figure~\ref{fig:stage6_classifier} shows progressively clearer separation between interacting and non-interacting pairs after dual-branch encoding and serial cross-attention, consistent with increasingly discriminative intermediate representations.

\begin{figure}[!htbp]
	\centering
	\includegraphics[width=0.94\columnwidth]{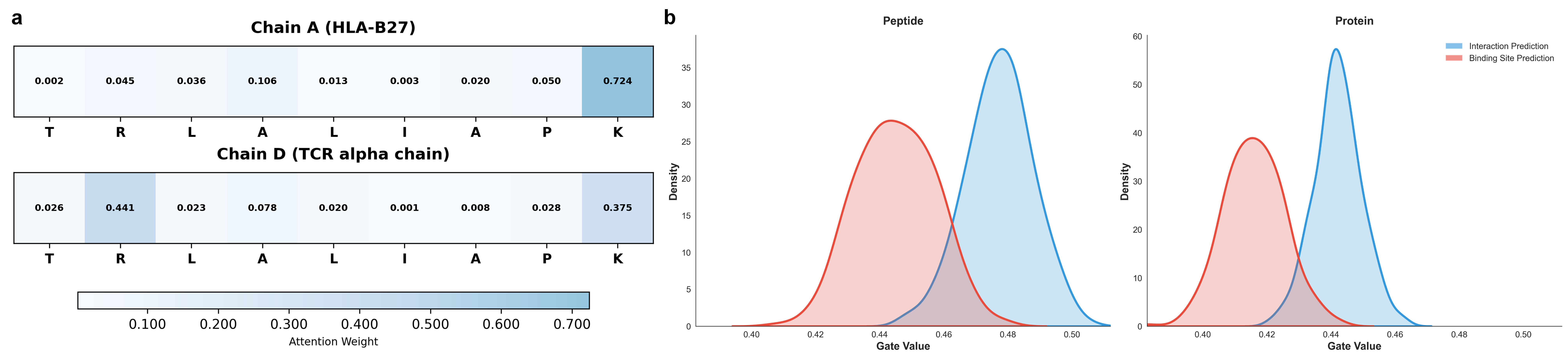}
	\caption{Interpretability analysis of the prediction model. (A) Residue-wise attention weights for the same peptide under different protein contexts, showing partner-dependent attention redistribution. (B) Distributions of gated-fusion activations for interaction prediction and binding-site prediction on the peptide and protein sides. Interaction prediction shows higher gate values, consistent with stronger use of global contextual information.}
	\label{fig:predic_explain}
\end{figure}

We next examined how this partner dependence appears at the residue and feature-integration levels. In the 7N2N case study, bidirectional attention shifted across complexes, indicating that the same peptide can redistribute attention under different protein contexts. Gated-fusion activations were also higher for interaction prediction than for binding-site prediction (Figure~\ref{fig:predic_explain}), consistent with stronger reliance on global contextual information at the sequence level.

\begin{figure}[!htbp]
	\centering
	\includegraphics[width=0.78\columnwidth]{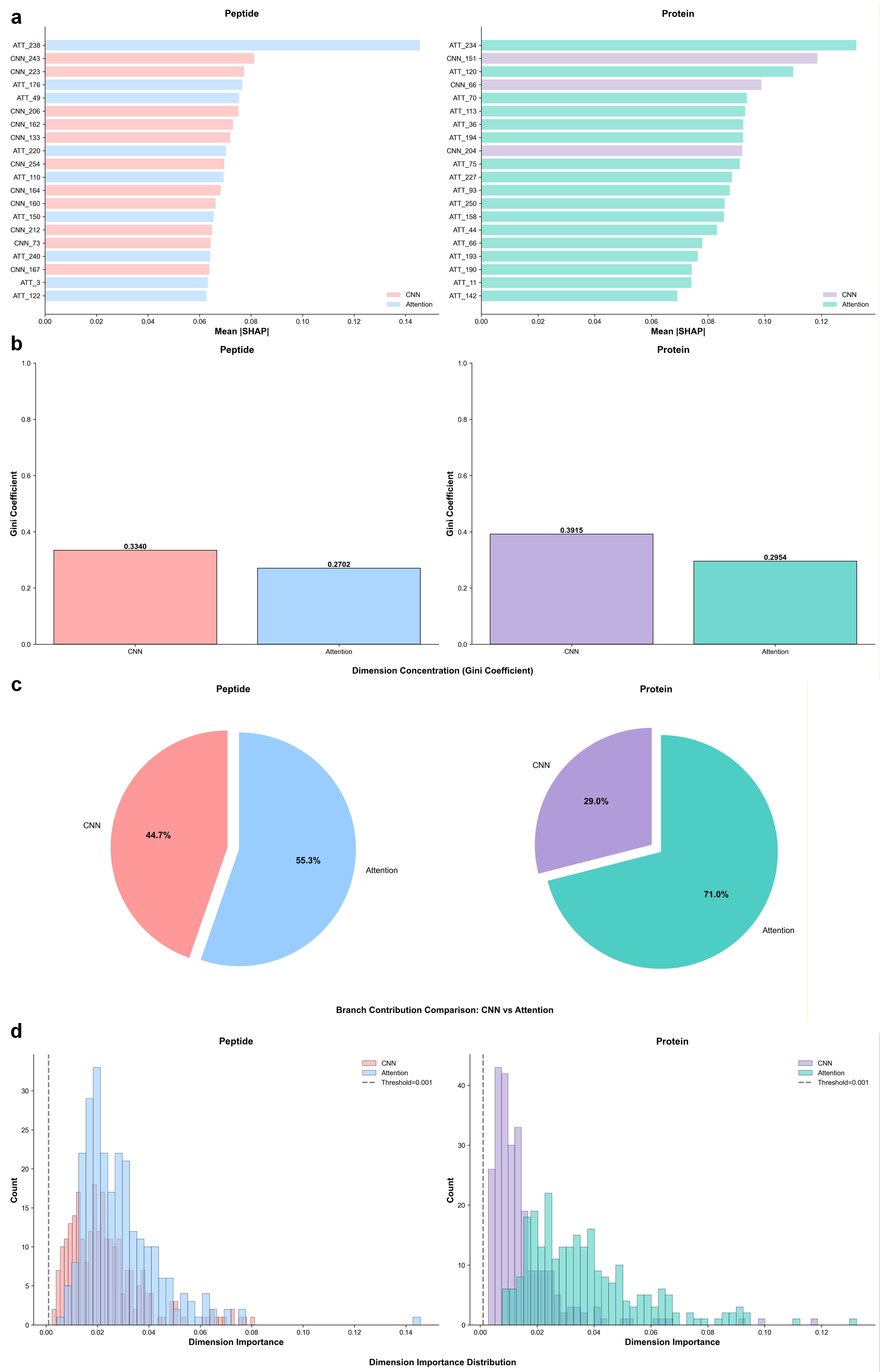}
	\caption{SHAP-based analysis of dual-channel feature contributions. Panels show feature importance magnitudes, branch-level concentration, aggregate contribution ratios, and per-dimension importance distributions. Attention dominates the overall contribution profile, while CNN features provide complementary local information.}
	\label{fig:shap_analyse}
\end{figure}

SHAP analysis further clarifies the fused features. In the 1024-dimensional representation, attention contributes more than CNN features (667.03 versus 389.39), with the protein-side attention branch contributing the most (373.86). Protein CNN has the highest Gini coefficient (0.3915), whereas peptide attention is the most evenly distributed (0.2702) (Figure~\ref{fig:shap_analyse}).

\subsection{Controlled Benchmarking of Target-Conditioned Generation}
To assess whether persistent protein conditioning improves peptide generation when sequence length is held constant, we evaluated TC-PepGen under a controlled target-specific benchmark. All methods were given the native binder length at test time, matching PepMLM and RFdiffusion.

We assessed the generated sequences with Chai-1, ESMFold, and AlphaFold 3, defining a hit as a generated binder whose ipTM exceeded that of the corresponding native test peptide. Across all three evaluators, TC-PepGen consistently outperformed PepMLM and RFdiffusion, indicating stronger predicted structural plausibility under this controlled benchmark.

\begin{figure}[H]
	\centering
	\includegraphics[width=0.74\textwidth,height=0.63\textheight,keepaspectratio]{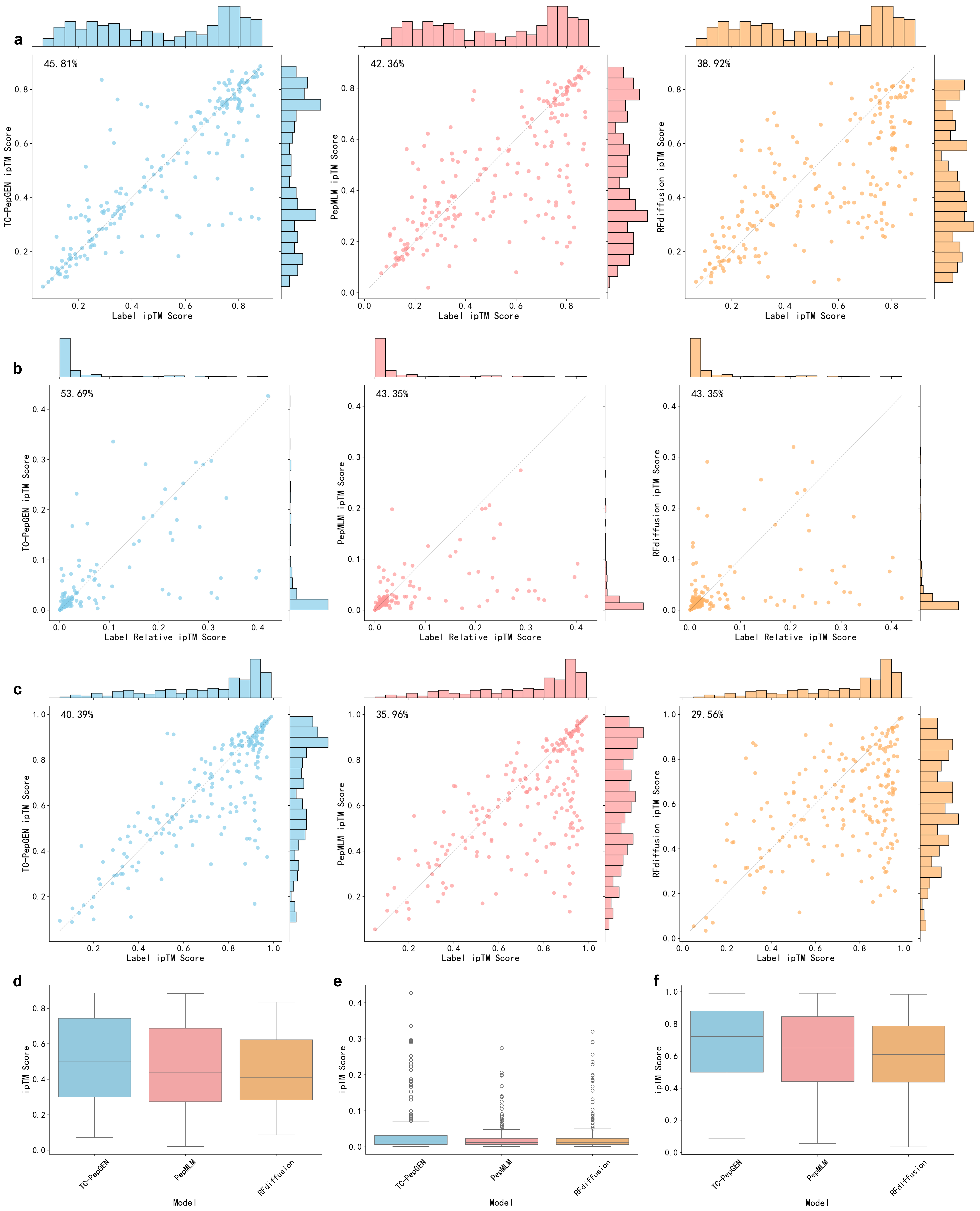}
	\caption{Extended comparison of generative model performance across TC-PepGen and baseline generators. (A) Scatter plots of Chai-1 ipTM scores for generated peptides versus native test peptides, with marginal histograms and the percentage of generated sequences that exceed the native template. (B) Scatter plots of relative ipTM scores under ESMFold using the same comparison scheme. (C) Scatter plots of AlphaFold 3 ipTM scores for generated peptides versus native test peptides. (D) Box-plot summary of Chai-1 ipTM distributions across TC-PepGen, PepMLM, and RFdiffusion. (E) Box-plot summary of relative ipTM distributions under ESMFold. (F) Box-plot summary of AlphaFold 3 ipTM distributions. Across all three evaluators, TC-PepGen shows a larger proportion of generated peptides surpassing the native templates and a consistently stronger score distribution than the baseline generators.}
	\label{fig:gen_compare}
\end{figure}

With Chai-1, 45.81\% of TC-PepGen sequences exceeded the native templates, compared with 42.36\% for PepMLM and 38.92\% for RFdiffusion. Under ESMFold, using a relative ipTM score derived from predicted alignment error, TC-PepGen reached 53.69\%, whereas PepMLM and RFdiffusion both reached 43.35\%. AlphaFold 3 showed the same ordering, with 40.39\% for TC-PepGen, 35.96\% for PepMLM, and 29.56\% for RFdiffusion. Figure~\ref{fig:gen_compare} provides the full score distributions.

\begin{figure}[!htbp]
	\centering
	\includegraphics[width=0.72\textwidth]{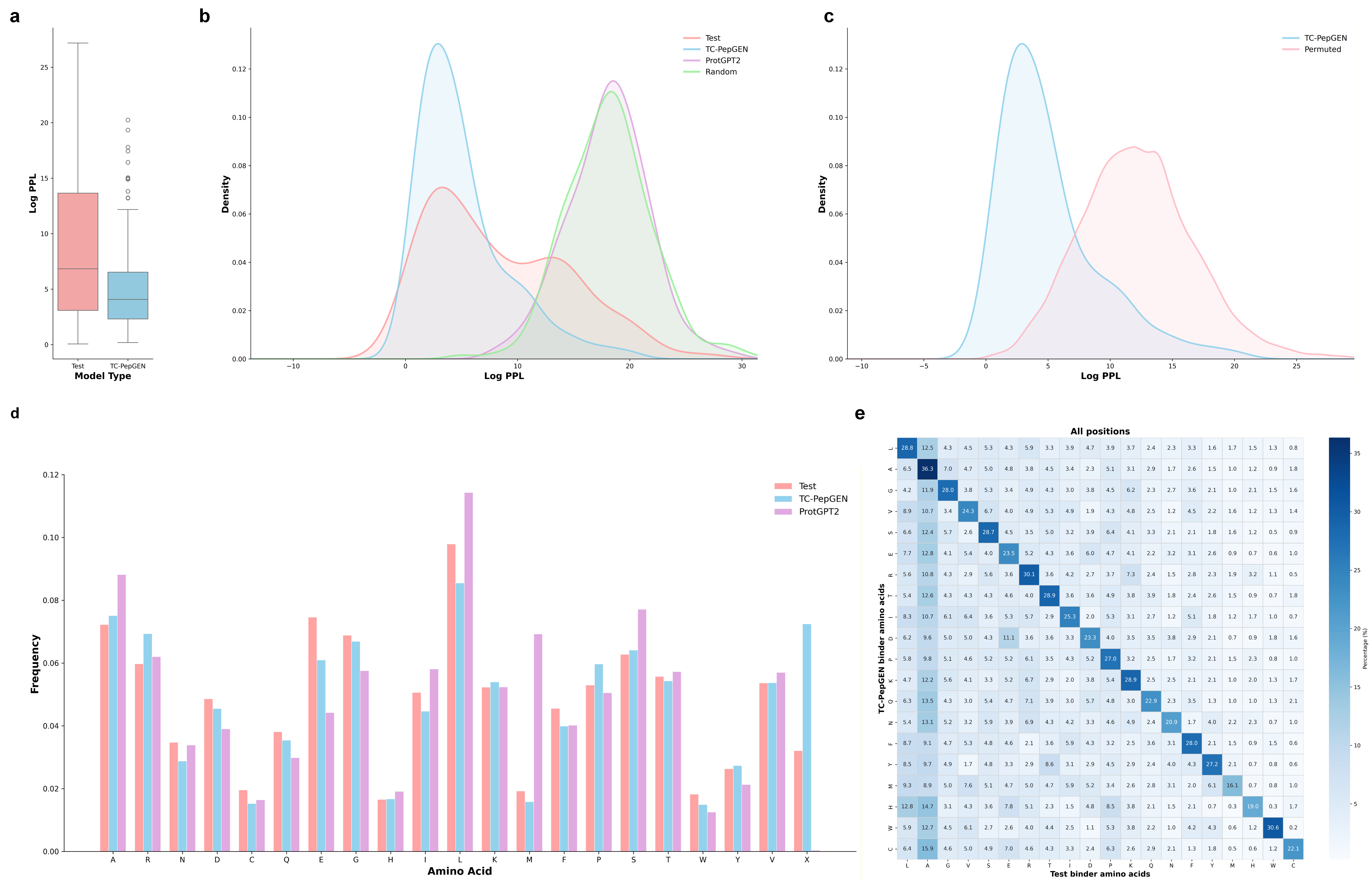}
	\caption{Sequence-level quality analysis of peptides generated by TC-PepGen. (A) Box-plot comparison of log perplexity (Log PPL) between native test peptides and generated peptides. (B) Log PPL distributions for native, TC-PepGen, ProtGPT2, and random sequences. (C) Comparison between TC-PepGen sequences and permuted controls. (D) Amino-acid frequency distributions. (E) Amino-acid confusion matrix between generated and test binders. TC-PepGen more closely matches native binders than baseline or control sequences while retaining diversity.}
	\label{fig:seq_analyse}
\end{figure}

Beyond benchmark hit rates, we next examined whether TC-PepGen also preserves native-like sequence statistics and amino-acid usage patterns. Figure~\ref{fig:seq_analyse} summarizes these sequence-level analyses: generated peptides remain closer to native binders than ProtGPT2, random, or permuted controls in perplexity, composition, and confusion-matrix structure, while still retaining diversity.

\FloatBarrier

\subsection{Retention of Target-Conditioned Signal without Length Constraints}

We then asked whether the target-conditioned signal persists when the model no longer has access to the true peptide length. Even under this unconstrained condition, 27.09\% of generated peptides achieved AlphaFold 3 interface prediction scores above the native templates, suggesting that the controlled-benchmark gains are not explained solely by access to the true length.

\begin{figure}[!htbp]
	\centering
	\includegraphics[width=0.82\textwidth,height=0.42\textheight,keepaspectratio]{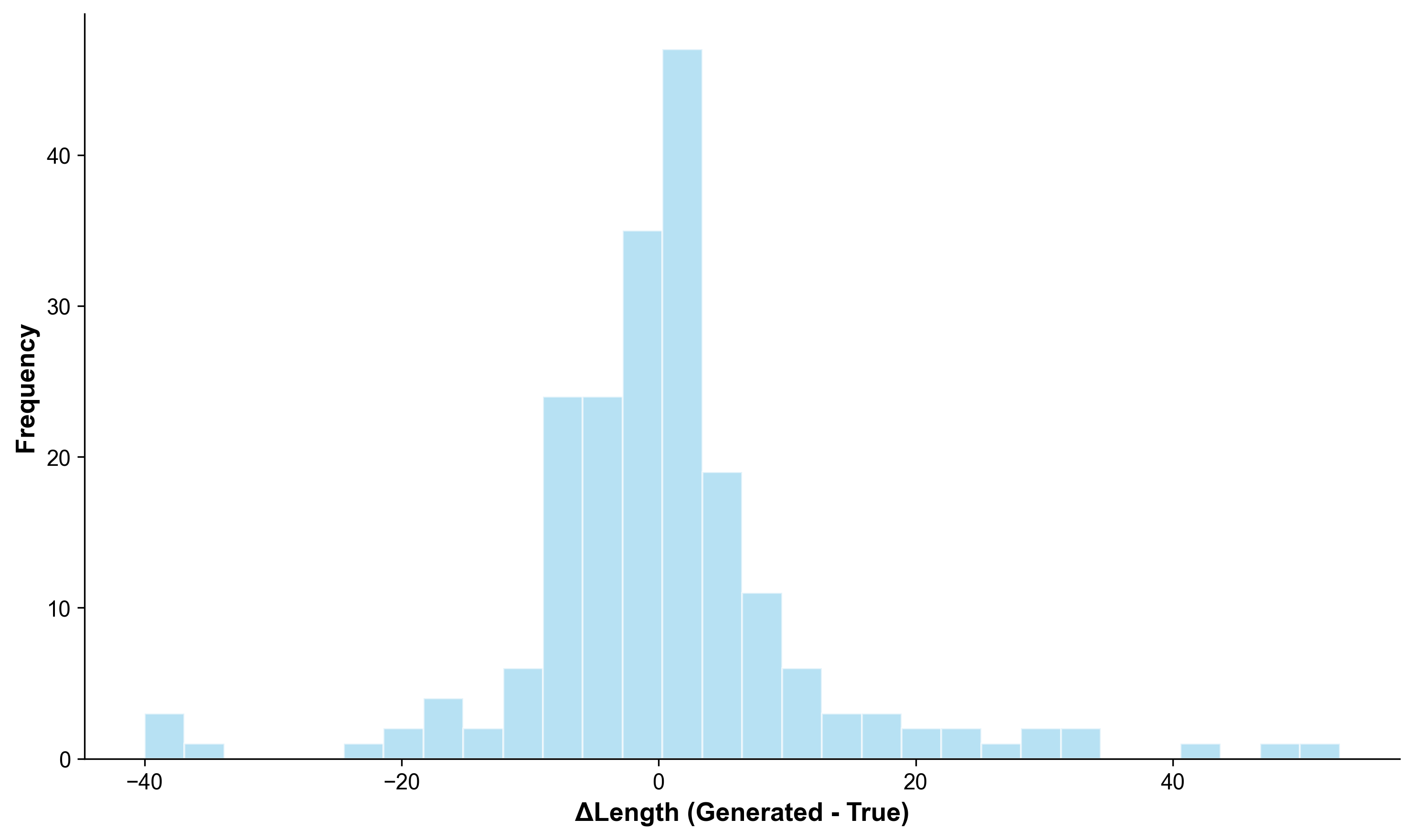}
	\caption{Length distribution under unconstrained natural generation. Histogram of peptide-length deviation, defined as generated length minus reference binder length. Most generated sequences remain close to the reference length, with the distribution concentrated near zero and slightly shifted toward modest positive deviations. Although both shorter and longer peptides are observed, the majority of samples stay within a biologically plausible range rather than diverging to unrealistic lengths.}
	\label{fig:length_deviation}
\end{figure}

To interpret this drop, we quantified the deviation between generated and reference peptide lengths as $\Delta \text{Length} = \text{Generated Length} - \text{True Length}$ (Figure~\ref{fig:length_deviation}). The distribution peaks between 0 and $+5$, indicating that TC-PepGen still generates peptides within a biologically plausible range and that part of the remaining gap likely reflects imperfect length recovery rather than loss of target-conditioned signal.

\subsection{Sequence Evidence for Target-Conditioned Generation}

To examine whether TC-PepGen learns a target-conditioned peptide prior rather than behaving as an unconditional peptide language model, we analyzed sequence-level statistics together with cross-attention behavior.

\begin{figure}[!htbp]
	\centering
	\includegraphics[width=0.82\textwidth,height=0.48\textheight,keepaspectratio]{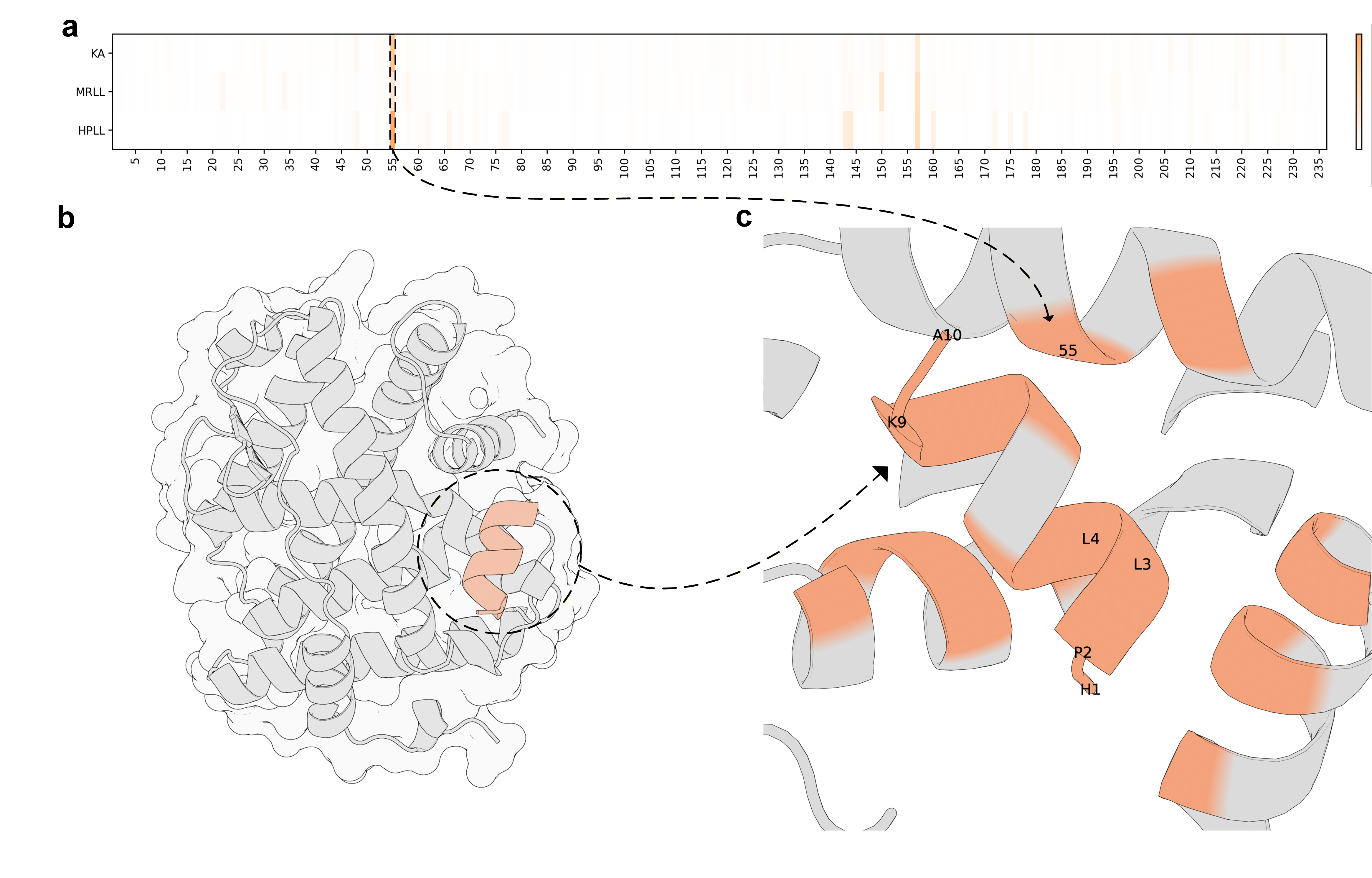}
	\caption{Cross-attention-based analysis of the TC-PepGen generative mechanism in the held-out 1ZKY complex. Panel (A) shows the residue-level cross-attention map from the final decoder layer between the target protein and the generated peptide, with a prominent attention peak near residue 55 indicating that the decoder focuses on a restricted subset of protein residues during generation. Panels (B) and (C) show that the highly attended regions cluster near the predicted binding groove and shift dynamically across decoding steps, indicating that conditioning evolves with the generated sequence prefix.}
	\label{fig:cross_attention_analysis}
\end{figure}

Cross-attention visualization provides complementary mechanistic support for this interpretation. In a representative held-out complex (PDB 1ZKY; Figure~\ref{fig:cross_attention_analysis}), final-layer decoder attention was concentrated on restricted target-protein regions rather than distributed uniformly across the sequence. A prominent peak appeared near residue 55, and the attended region shifted across decoding steps, indicating dynamic conditioning as the peptide prefix grew. When mapped onto the AlphaFold 3 complex model, high-attention regions clustered near the predicted binding groove, supporting the view that TC-PepGen repeatedly accesses target-relevant protein context during autoregressive generation.

\FloatBarrier

\subsection{Screening-Oriented Integration in Design Scenarios}

To illustrate the practical utility of the integrated framework in screening-oriented design scenarios, the following case studies show how target-conditioned generation, prediction-guided ranking, and structure-based follow-up can be combined to support candidate triage. In these applications, the framework enriches higher-ranked candidates and remains useful under stringent low-homology conditions.

\subsubsection{Prediction-Guided Ranking in a GSK-3 beta Screening Example}

\begin{figure}[H]
	\centering
	\includegraphics[width=0.42\textwidth,height=0.48\textheight,keepaspectratio]{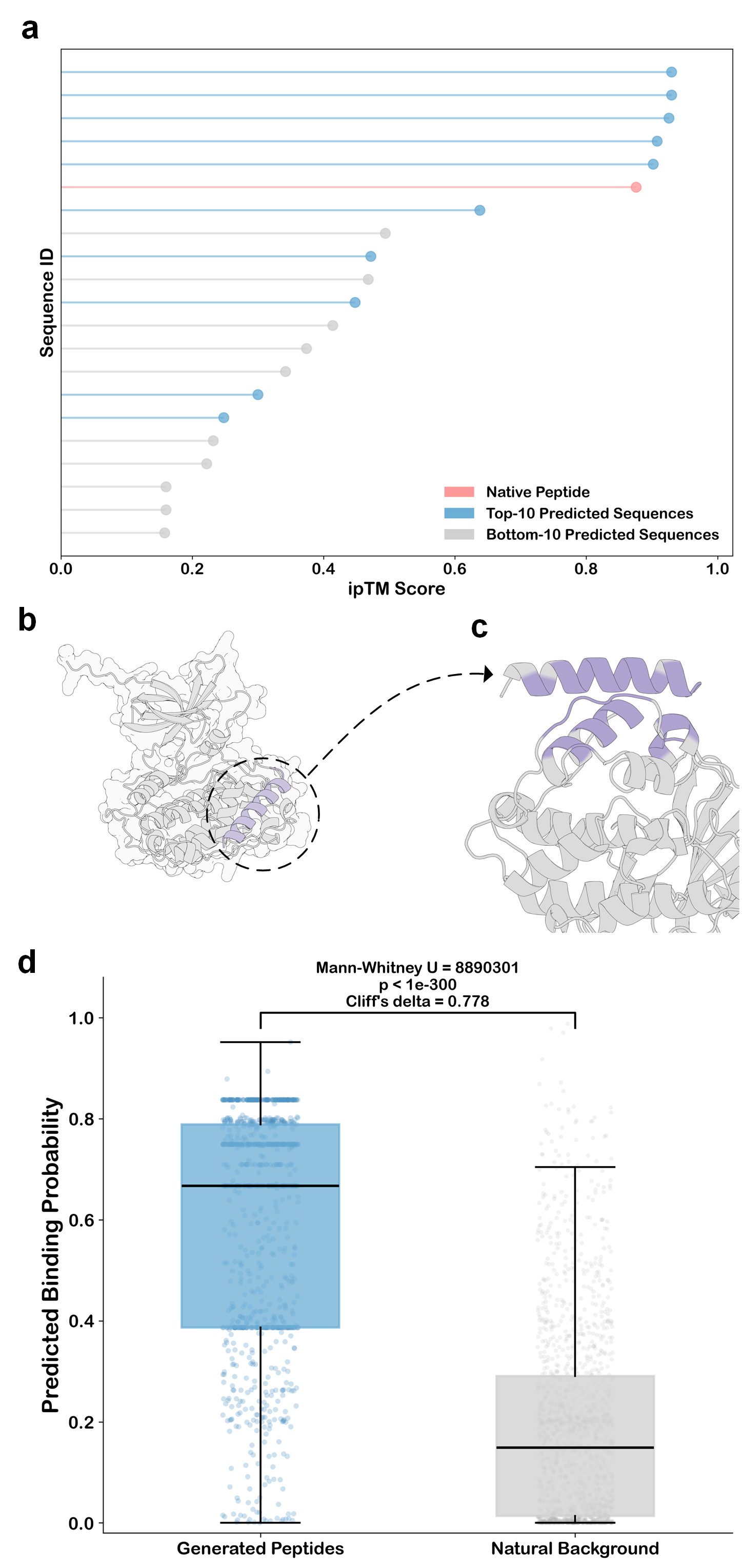}
	\caption{Prediction-guided candidate triage on GSK-3 beta. Panels show AlphaFold 3 scores for native, top-ranked, and bottom-ranked candidates, a representative top-ranked complex structure with interface zoom-in, and the predicted binding-probability distributions of generated candidates versus a natural peptide background.}
	\label{fig:closed_loop_validation}
\end{figure}

To illustrate prediction-guided candidate triage in a practical target setting, the integrated framework enriched higher-ranked GSK-3 beta candidates for more favorable structure-based scores. We generated 1000 candidate peptides for GSK-3 beta with TC-PepGen, ranked them with ConGA-PepPI, and evaluated the top and bottom 10 candidates together with the native peptide by AlphaFold 3 using ipTM. Figure~\ref{fig:closed_loop_validation}A--C shows that the top-ranked sequences consistently receive more favorable structure-based scores than the bottom-ranked set, and some also exceed the native peptide. The best generated peptide occupies the GSK-3 beta binding groove and forms a plausible interface.

To move beyond structural inspection of only the highest- and lowest-ranked candidates, we next compared the full predicted binding-probability distribution of generated peptides against a natural peptide background. We retained the original predicted probabilities for all 1000 GSK-3 beta-generated candidates and constructed a control set of 10000 UniProt-derived natural peptides. Because all generated candidates in this experiment were 21-mers, the control peptides were also sampled as 21-mers. Both groups were paired with the same GSK-3 beta target and scored with the same prediction model. As shown in Figure~\ref{fig:closed_loop_validation}D, generated peptides received substantially higher predicted binding probabilities than the natural background (mean 0.5927 versus 0.1910; median 0.6671 versus 0.1492; Mann--Whitney $U = 8{,}890{,}301$, $p < 10^{-300}$; Cliff's $\delta = 0.778$). The common-language effect size was 0.889, meaning that a randomly selected generated peptide would outscore a randomly selected natural background peptide in 88.9\% of pairwise comparisons. These results show enrichment across the overall generated-peptide distribution, not only among extreme high-scoring candidates.

\subsubsection{Key Residue Identification for Directed Optimization}

We next used in silico alanine scanning as an application case to examine whether ConGA-PepPI can prioritize sequence positions for downstream peptide optimization. The analysis was carried out on HLA-B27 in PDB structure 1FG2, which was not included in training, so this case also provides a qualitative check of generalization beyond memorized examples.

We first generated an 11-mer peptide, EGPRNQDWLIW, for the target protein and then constructed a full single-residue substitution panel. Each position was mutated to alanine, except that native alanine residues were replaced with glycine to preserve a non-trivial perturbation at every site. ConGA-PepPI was then used to score the wild-type peptide together with all 11 single-site mutants, and residue sensitivity was quantified as the relative decrease in predicted binding probability, $(P_{WT} - P_{Mut}) / P_{WT}$, where $P_{WT}$ and $P_{Mut}$ denote the predicted binding probabilities of the wild-type peptide and its single-site mutant, respectively. Larger sensitivity values indicate positions whose mutation is predicted to cause a stronger loss of interaction probability and therefore may represent candidate hotspots for optimization or preservation.

\begin{figure}[H]
	\centering
	\includegraphics[width=0.60\textwidth,height=0.42\textheight,keepaspectratio]{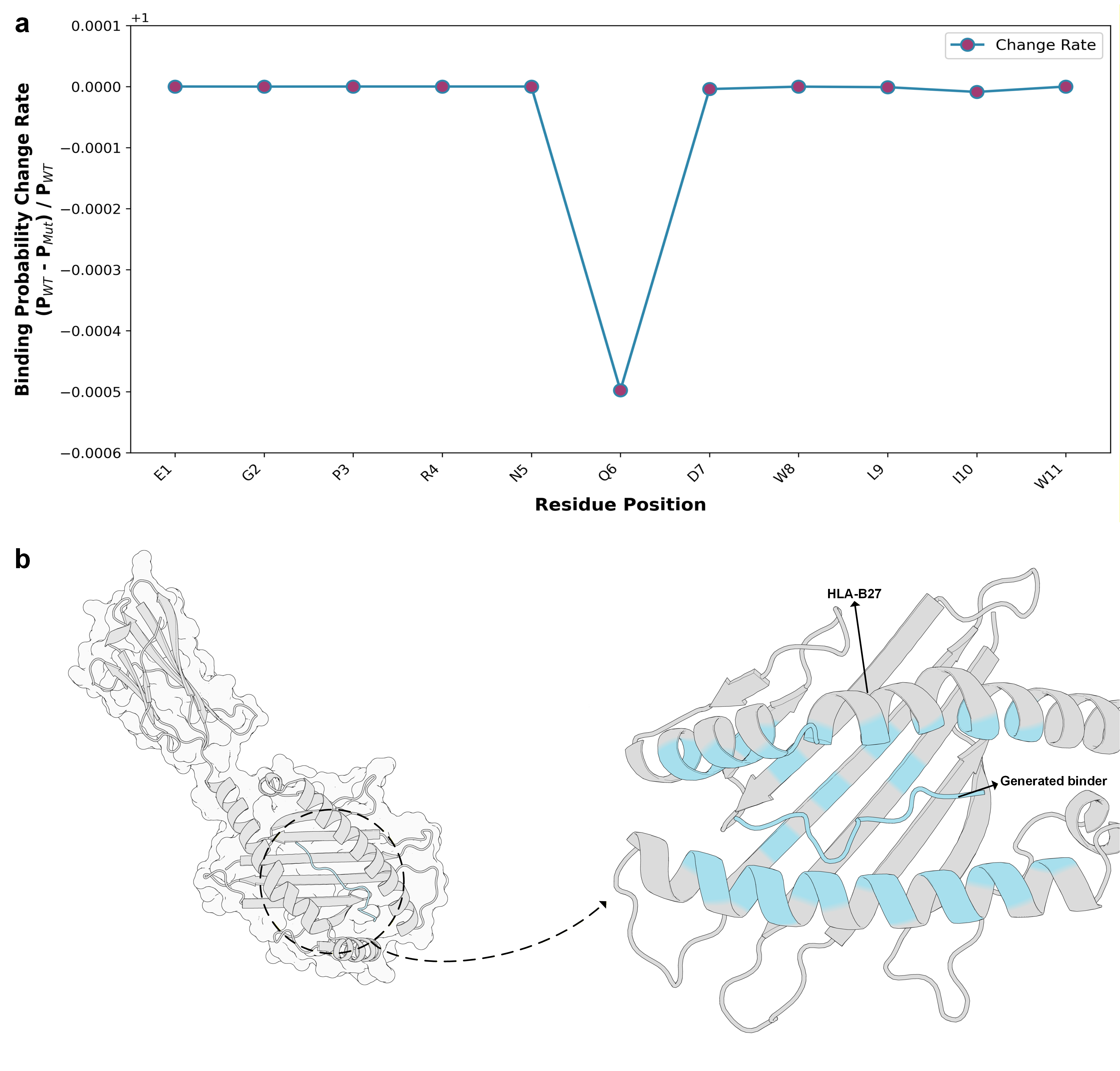}
	\caption{In silico alanine-scanning analysis of generated peptide candidates. Panel (A) shows residue-level sensitivity scores derived from single-site alanine substitution, highlighting positions whose mutation leads to the largest reduction in predicted interaction probability. Panel (B) shows the corresponding predicted complex structure, in which the most sensitive residues are located near the modeled peptide--protein interface. The agreement between high-sensitivity sequence positions and the predicted structural contact region supports the consistency between sequence-based sensitivity analysis and structure-based interface localization.}
	\label{fig:silico_alanine}
\end{figure}

Figure~\ref{fig:silico_alanine}A shows that sensitivity is distributed across most sequence positions rather than being concentrated at a single residue, suggesting that the generated peptide contains a broad interaction pattern with multiple potentially important contacts. To examine whether the sequence-based sensitivity profile is structurally plausible, we folded the target-peptide complex with AlphaFold 3. The resulting model places the peptide in the HLA-B27 binding groove, and the most sensitive residues map near the predicted interface (Figure~\ref{fig:silico_alanine}B). Although this analysis is still computational and does not establish experimental binding energetics, it illustrates how ConGA-PepPI can be used as a front-end tool for identifying residues to preserve, mutate, or prioritize in follow-up design cycles.

\FloatBarrier

\subsubsection{A Low-Homology MDM2 Design Example}

To examine applicability beyond familiar sequence space, we next considered a stringent low-homology MDM2 setting. We removed all training samples with more than 60\% similarity on either the protein or peptide side to construct this scenario.

On the prediction side, the framework retained useful sensitivity to known MDM2 binders under this low-homology constraint. We evaluated MDM2 together with its known ligand pMI (LTFEHYWAQLTS) and four active analogs from the same motif family, obtaining predicted binding probabilities from 0.942 to 1.000.

On the generation side, the framework also supported candidate nomination beyond known sequence space. We generated 100 fixed-length candidate peptides for MDM2 with TC-PepGen, ranked them with ConGA-PepPI, and selected the highest-scoring sequence for structural evaluation. The selected peptide was the novel 14-mer GSRVTEQEIAMLQN. Under this stringent low-homology setting, AlphaFold 3 assigned this peptide ipTM = 0.84 (pTM = 0.76), compared with ipTM = 0.87 (pTM = 0.74) for pMI, despite near-zero sequence identity between the generated peptide and pMI.

\section{Discussion and Conclusion}\label{sec:discussion}

On the prediction side, the results indicate that PepPI recognition benefits from modeling partner asymmetry and partner-dependent context rather than simply increasing model complexity, and that sequence-level interaction learning provides a useful foundation for residue localization. On the generation side, the results support the value of preserving target information throughout autoregressive decoding.

Future work will extend evaluation to broader peptide--protein benchmarks and strengthen downstream validation in more realistic early-stage screening settings, including prospective experimental assessment. Coupling target-conditioned generation with iterative prediction-guided ranking may further improve candidate triage and optimization.

\section*{Acknowledgements}

The authors thank their colleagues for helpful discussions.

\bibliographystyle{unsrtnat}
\nocite{*}
\FloatBarrier
\bibliography{bibliography}

\begin{thebibliography}{30}
\providecommand{\natexlab}[1]{#1}
\providecommand{\url}[1]{\texttt{#1}}
\expandafter\ifx\csname urlstyle\endcsname\relax
  \providecommand{\doi}[1]{doi: #1}\else
  \providecommand{\doi}{doi: \begingroup \urlstyle{rm}\Url}\fi

\bibitem[Liu et~al.(2018)Liu, Jobichen, Chia, Chan, others, and Tenen]{liu2018sall4}
B.~H. Liu, C.~Jobichen, C.~S.~B. Chia, T.~H.~M. Chan, others, and D.~G. Tenen.
\newblock Targeting cancer addiction for {SALL4} by shifting its transcriptome with a pharmacologic peptide.
\newblock \emph{Proceedings of the National Academy of Sciences of the United States of America}, 115\penalty0 (30):\penalty0 E7119--E7128, 2018.

\bibitem[Xu et~al.(2024)Xu, Fan, He, Xia, and Zhang]{xu2024lysineStapled}
L.~Xu, X.~Fan, Y.~He, X.~Xia, and J.~Zhang.
\newblock Design, synthesis, and biological evaluation of lysine-stapled peptide inhibitors of p53-{MDM2}/{MDMX} interactions with potent antitumor activity in vivo.
\newblock \emph{Journal of Medicinal Chemistry}, 67:\penalty0 17893--17904, 2024.

\bibitem[Kurcinski et~al.(2015)Kurcinski, Jamroz, Blaszczyk, Kolinski, and Kmiecik]{kurcinski2015cabsdock}
M.~Kurcinski, M.~Jamroz, M.~Blaszczyk, A.~Kolinski, and S.~Kmiecik.
\newblock {CABS-dock} web server for the flexible docking of peptides to proteins without prior knowledge of the binding site.
\newblock \emph{Nucleic Acids Research}, 43\penalty0 (W1):\penalty0 W419--W424, 2015.
\newblock \doi{10.1093/nar/gkv456}.

\bibitem[Xu et~al.(2018)Xu, Yan, and Zou]{xu2018mdockpep}
X.~Xu, C.~Yan, and X.~Zou.
\newblock {MDockPeP}: an ab-initio protein-peptide docking server.
\newblock \emph{Journal of Computational Chemistry}, 39\penalty0 (28):\penalty0 2409--2413, 2018.
\newblock \doi{10.1002/jcc.25555}.

\bibitem[Zhang and Sanner(2019)]{zhang2019autodockcrankpep}
Y.~Zhang and M.~F. Sanner.
\newblock {AutoDock CrankPep}: combining folding and docking to predict protein-peptide complexes.
\newblock \emph{Bioinformatics}, 35\penalty0 (24):\penalty0 5121--5127, 2019.
\newblock \doi{10.1093/bioinformatics/btz459}.

\bibitem[Lee et~al.(2015)Lee, Heo, Lee, and Seok]{lee2015galaxypepdock}
H.~Lee, L.~Heo, M.~S. Lee, and C.~Seok.
\newblock {GalaxyPepDock}: a protein-peptide docking tool based on interaction similarity and energy optimization.
\newblock \emph{Nucleic Acids Research}, 43\penalty0 (W1):\penalty0 W431--W435, 2015.
\newblock \doi{10.1093/nar/gkv495}.

\bibitem[Lei et~al.(2021)Lei, Li, Liu, Wan, Wang, Zhong, et~al.]{lei2021camp}
Y.~Lei, S.~Li, H.~Liu, F.~Wan, X.~Wang, L.~Zhong, et~al.
\newblock A deep-learning framework for multi-level peptide-protein interaction prediction.
\newblock \emph{Nature Communications}, 12\penalty0 (1):\penalty0 5465, 2021.
\newblock \doi{10.1038/s41467-021-25772-4}.

\bibitem[Wu et~al.(2022)Wu, Gao, Zeng, Zhang, and Li]{wu2022bridgedpi}
Y.~Wu, M.~Gao, M.~Zeng, J.~Zhang, and M.~Li.
\newblock {BridgeDPI}: a novel graph neural network for predicting drug-protein interactions.
\newblock \emph{Bioinformatics}, 38\penalty0 (9):\penalty0 2571--2578, 2022.
\newblock \doi{10.1093/bioinformatics/btac140}.

\bibitem[Wang et~al.(2024)]{wang2024deeppeppi}
Z.~Wang et~al.
\newblock {DeepPepPI}: a deep cross-dependent framework with information sharing mechanism for predicting plant peptide-protein interactions.
\newblock \emph{Expert Systems with Applications}, 252:\penalty0 124168, 2024.

\bibitem[Chen et~al.(2025{\natexlab{a}})Chen, Yan, Li, and Liu]{chen2025plpa}
S.~Chen, K.~Yan, X.~Li, and B.~Liu.
\newblock Protein language pragmatic analysis and progressive transfer learning for profiling peptide-protein interactions.
\newblock \emph{IEEE Transactions on Neural Networks and Learning Systems}, 36\penalty0 (8):\penalty0 15385--15399, 2025{\natexlab{a}}.

\bibitem[Chen et~al.(2025{\natexlab{b}})Chen, Quinn, Dumas, Peng, Hong, Lopez-Gonzalez, Mestre, Watson, Vincoff, Zhao, Wu, Stavrand, Schaepers-Cheu, Wang, Srijay, Monticello, Vure, Pulugurta, Pertsemlidis, and Chatterjee]{chen2025targetSequence}
L.~T. Chen, Z.~Quinn, M.~Dumas, C.~Peng, L.~Hong, M.~Lopez-Gonzalez, A.~Mestre, R.~Watson, S.~Vincoff, L.~Zhao, J.~Wu, A.~Stavrand, M.~Schaepers-Cheu, T.~Z. Wang, D.~Srijay, C.~Monticello, P.~Vure, R.~Pulugurta, S.~Pertsemlidis, and P.~Chatterjee.
\newblock Target sequence-conditioned design of peptide binders using masked language modeling.
\newblock \emph{Nature Biotechnology}, 2025{\natexlab{b}}.
\newblock \doi{10.1038/s41587-025-02761-2}.

\bibitem[Jin et~al.(2024)Jin, Chen, Yu, Jiang, Chen, Yan, others, and Wang]{jin2024tpeppro}
X.~Jin, Z.~Chen, D.~Yu, Q.~Jiang, Z.~Chen, B.~Yan, others, and J.~Wang.
\newblock {TPepPro}: a deep learning model for predicting peptide-protein interactions.
\newblock \emph{Bioinformatics}, 41\penalty0 (1):\penalty0 btae708, 2024.
\newblock \doi{10.1093/bioinformatics/btae708}.

\bibitem[Zeng et~al.(2025)Zeng, Liu, Yu, Han, and Liu]{zeng2025collaborative}
L.~Zeng, Y.~Liu, Z.~Yu, G.~Han, and Y.~Liu.
\newblock Collaborative learning macroscopic binding trends and microscopic residue interactions to predict peptide-protein interactions.
\newblock \emph{IEEE Journal of Biomedical and Health Informatics}, 2025.

\bibitem[Chen et~al.(2025{\natexlab{c}})Chen, Quinn, Zhang, and Chatterjee]{chen2025moppitv3}
T.~Chen, Z.~Quinn, Y.~Zhang, and P.~Chatterjee.
\newblock {moPPIt-v3}: Motif-specific peptides generated via multi-objective-guided discrete flow matching, 2025{\natexlab{c}}.
\newblock Preprint.

\bibitem[Chen et~al.(2025{\natexlab{d}})Chen, Chen, Lin, Chen, and Cheng]{chen2025geopep}
D.~Chen, Y.~Chen, T.~Lin, S.~Chen, and X.~Cheng.
\newblock {GeoPep}: A geometry-aware masked language model for protein-peptide binding site prediction.
\newblock \emph{arXiv}, 2025{\natexlab{d}}.

\bibitem[Zhai et~al.(2025)Zhai, Wang, Tang, Zhong, Xu, Liu, others, and Lu]{zhai2025peptideLm}
J.~Zhai, Z.~Wang, C.~Tang, H.~Zhong, Z.~Xu, Y.~Liu, others, and T.~Lu.
\newblock A general language model for peptide function identification.
\newblock \emph{arXiv}, 2025.

\bibitem[Wang et~al.(2023)Wang, Li, Ming, Wu, Fang, Huang, Lin, Liu, Kuang, Zhao, Huang, Feng, Guo, Yang, Guo, Zhang, Chen, Liu, Zhu, and Pei]{wang2023nurd}
B.~Wang, C.~Li, J.~Ming, L.~Wu, S.~Fang, Y.~Huang, L.~Lin, H.~Liu, J.~Kuang, C.~Zhao, X.~Huang, H.~Feng, J.~Guo, X.~Yang, L.~Guo, X.~Zhang, J.~Chen, J.~Liu, P.~Zhu, and D.~Pei.
\newblock The {NuRD} complex cooperates with {SALL4} to orchestrate reprogramming.
\newblock \emph{Nature Communications}, 14:\penalty0 2846, 2023.

\bibitem[Rew and Sun(2014)]{rew2014amg232}
Y.~Rew and D.~Sun.
\newblock Discovery of a small molecule {MDM2} inhibitor ({AMG} 232) for treating cancer.
\newblock \emph{Journal of Medicinal Chemistry}, 57:\penalty0 6332--6341, 2014.

\bibitem[Bai et~al.(2023)Bai, Miljkovi{\'c}, John, and Lu]{bai2023drugban}
P.~Bai, F.~Miljkovi{\'c}, B.~John, and H.~Lu.
\newblock Interpretable bilinear attention network with domain adaptation improves drug-target prediction.
\newblock \emph{Nature Machine Intelligence}, 5\penalty0 (2):\penalty0 126--136, 2023.

\bibitem[{\"O}zt{\"u}rk et~al.(2018){\"O}zt{\"u}rk, {\"O}zg{\"u}r, and Ozkirimli]{ozturk2018deepdta}
H.~{\"O}zt{\"u}rk, A.~{\"O}zg{\"u}r, and E.~Ozkirimli.
\newblock {DeepDTA}: deep drug-target binding affinity prediction.
\newblock \emph{Bioinformatics}, 34\penalty0 (17):\penalty0 i821--i829, 2018.
\newblock \doi{10.1093/bioinformatics/bty593}.

\bibitem[Tsubaki et~al.(2019)Tsubaki, Tomii, and Sese]{tsubaki2019cpi}
M.~Tsubaki, K.~Tomii, and J.~Sese.
\newblock Compound-protein interaction prediction with end-to-end learning of neural networks for graphs and sequences.
\newblock \emph{Bioinformatics}, 35\penalty0 (2):\penalty0 309--318, 2019.
\newblock \doi{10.1093/bioinformatics/bty535}.

\bibitem[Abramson et~al.(2024)]{abramson2024alphafold3}
J.~Abramson et~al.
\newblock Accurate structure prediction of biomolecular interactions with {AlphaFold} 3.
\newblock \emph{Nature}, 630\penalty0 (8016):\penalty0 493--500, 2024.
\newblock \doi{10.1038/s41586-024-07487-w}.

\bibitem[Elnaggar et~al.(2021)Elnaggar, Heinzinger, Dallago, Rehawi, Wang, Jones, et~al.]{elnaggar2021prottrans}
A.~Elnaggar, M.~Heinzinger, C.~Dallago, G.~Rehawi, Y.~Wang, L.~Jones, et~al.
\newblock {ProtTrans}: toward understanding the language of life through self-supervised learning.
\newblock \emph{IEEE Transactions on Pattern Analysis and Machine Intelligence}, 44\penalty0 (10):\penalty0 7112--7127, 2021.
\newblock \doi{10.1109/TPAMI.2021.3095381}.

\bibitem[Abdin et~al.(2022)Abdin, Nim, Wen, and Kim]{abdin2022pepnn}
O.~Abdin, S.~Nim, H.~Wen, and P.~M. Kim.
\newblock {PepNN}: a deep attention model for the identification of peptide binding sites.
\newblock \emph{Communications Biology}, 5\penalty0 (1):\penalty0 503, 2022.
\newblock \doi{10.1038/s42003-022-03463-4}.

\bibitem[Das et~al.(2013)Das, Sharma, Kumar, Krishna, and Mathur]{das2013pepbind}
A.~A. Das, O.~P. Sharma, M.~S. Kumar, R.~Krishna, and P.~P. Mathur.
\newblock {PepBind}: a comprehensive database and computational tool for analysis of protein-peptide interactions.
\newblock \emph{Genomics, Proteomics \& Bioinformatics}, 11\penalty0 (4):\penalty0 241--246, 2013.
\newblock \doi{10.1016/j.gpb.2013.07.004}.

\bibitem[Wang et~al.(2022)Wang, Jin, Zou, Nakai, and Wei]{wang2022pepbcl}
R.~Wang, J.~Jin, Q.~Zou, K.~Nakai, and L.~Wei.
\newblock Predicting protein-peptide binding residues via interpretable deep learning.
\newblock \emph{Bioinformatics}, 38\penalty0 (13):\penalty0 3351--3360, 2022.
\newblock \doi{10.1093/bioinformatics/btac297}.

\bibitem[Huang et~al.(2024)]{huang2024pepca}
J.~Huang et~al.
\newblock {PepCA}: unveiling protein-peptide interaction sites with a multi-input neural network model.
\newblock \emph{iScience}, 27\penalty0 (10):\penalty0 110850, 2024.

\bibitem[Rao et~al.(2019)Rao, Bhattacharya, Thomas, Duan, Chen, Canny, Abbeel, and Song]{rao2019tape}
R.~Rao, N.~Bhattacharya, N.~Thomas, Y.~Duan, X.~Chen, J.~Canny, P.~Abbeel, and Y.~S. Song.
\newblock Evaluating protein transfer learning with {TAPE}.
\newblock In \emph{Advances in Neural Information Processing Systems}, volume~32, pages 9689--9701, 2019.

\bibitem[Lin et~al.(2023)Lin, Akin, Rao, Hie, Zhu, Lu, et~al.]{lin2023esm2}
Z.~Lin, H.~Akin, R.~Rao, B.~Hie, Z.~Zhu, W.~Lu, et~al.
\newblock Evolutionary-scale prediction of atomic-level protein structure with a language model.
\newblock \emph{Science}, 379\penalty0 (6637):\penalty0 1123--1130, 2023.
\newblock \doi{10.1126/science.ade2574}.

\bibitem[Xu et~al.(2025)Xu, Wang, Lu, and Zhai]{xu2025scmppi}
S.~Xu, Z.~Wang, T.~Lu, and J.~Zhai.
\newblock {SCMPPI}: Supervised contrastive multimodal framework for predicting protein-protein interactions.
\newblock \emph{arXiv}, 2025.

\end{thebibliography}

\end{document}